# Deep Learning Approach for Receipt Recognition

Anh Duc Le [1,2], Dung Van Pham[2], Tuan Anh Nguyen[2]
1) Center of Open Data in Humanities
Research Organization of Information and Systems
Tokyo, Japan
2) Deep Learning and Application, Vietnam

*Abstract*—Inspired by the recent successes of deep learning on Computer Vision and Natural Language Processing, we present a deep learning approach for recognizing scanned receipts. The recognition system has two main modules: text detection based on Connectionist Text Proposal Network and text recognition based on Attention-based Encoder-Decoder. We also proposed pre-processing to extract receipt area and OCR verification to ignore handwriting. The experiments on the dataset of the Robust Reading Challenge on Scanned Receipts OCR and Information Extraction 2019 demonstrate that the accuracies were improved by integrating the pre-processing and the OCR verification. Our recognition system achieved 71.9% of the F1 score for detection and recognition task.

## I. Introduction

Automatically identifying and extracting key texts from scanned structured and semi-structured receipts and saving them as structured data can serve as a very first step to enable a variety of applications. For example, the knowledge of consumer behavior can be extracted from those texts through data analytics. Pipelines for automating office paper works such as in accounting or taxation can also benefit from receipt information in the form of structured data. Although there are significant improvements of OCR in many tasks such as name card recognition, license plate recognition or handwritten text recognition by the recent advance in deep learning, receipt OCR is still challenging because of its higher requirements in accuracy to be practical. Moreover, in some cases, receipts with small size and handwriting are challenging for the recognition system. Figure 1 shows two examples of receipts with a small size and handwriting. For the above reasons, manual labeling is still widely used. Therefore, fast and reliable OCR needs to be developed in order to reduce or even eliminate manual works.

For text line detection, EAST [1] and Connectionist Text Proposal Network (CTPN) [2] have been proposed for detecting text in natural images. CTPN explores rich context information of an input image, making it powerful to detect horizontal text.

For text line recognition, Convolutional Recurrent Neural Network and Attention-based Encoder-Decoder (AED) have been achieved state-of-the-art for many problems such as Scene Text Recognition [3], handwritten recognition [4, 5, 6].

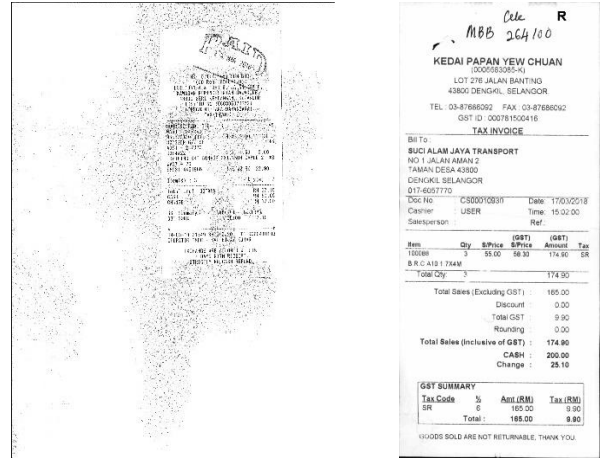

Fig. 1. Two examples of receipts with small size and handwriting.

In this paper, we presented a deep learning approach for text detection and recognition. We employed CTPN for text detection and AED for text recognition. Moreover, pre-processing and OCR verification were employed to extract small receipt and remove handwriting, respectively. Experimental results on the dataset from the Robust Reading Challenge on Scanned Receipts OCR and Information Extraction (SROIE 2019) demonstrate the robustness and efficiency of our system.

## II. Methodology

Inspired by recent successes of deep learning in many tasks such as computer vision, natural language processing, we present a deep learning approach for receipt recognition. The system has two stages based on deep learning: text detection and text recognition. We employ CTPN for text detection and AED for text recognition. The structures of text detection and recognition are shown in Figure 2 and 3.

### A. Text Detection

The original CTPN detects horizontally arranged text. The CTPN structure is basically similar to Faster R-CNN, but with the addition of the LSTM layer. The network model mainly consists of three parts: feature extraction by VGG16, bidirectional LSTM, and bounding box regression, as shown in Figure 2. The process of text detection is as follows:

(1) VGG16 extract feature from an input receipt. The size of the features is W×H×C.

(2) A sliding window on the output feature map is employed. In each row, the sequential windows connected to a Bi-directional LSTM (BLSTM).

(3) The BLSTM layer is connected to a 512D fully-connected layer. Then, the output layer predicts text/non-text scores, y-axis coordinates, and side-refinement offsets of 10.

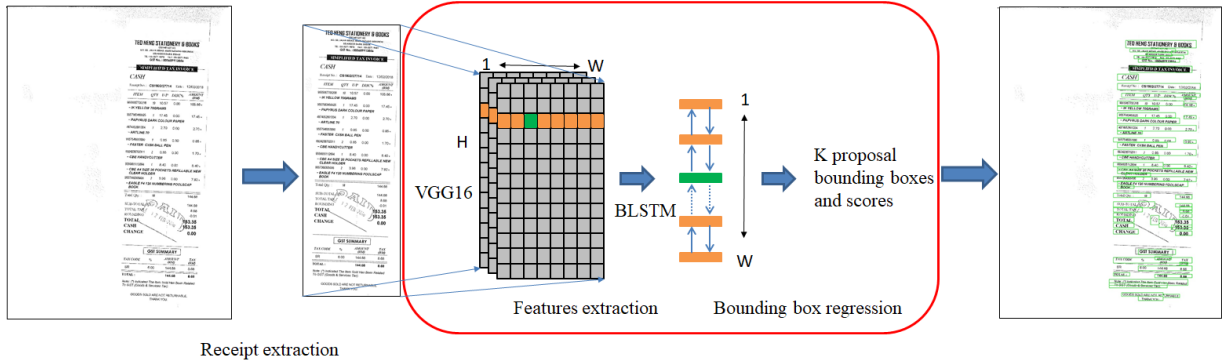

Fig. 2. Structure of CTPN for text detection.

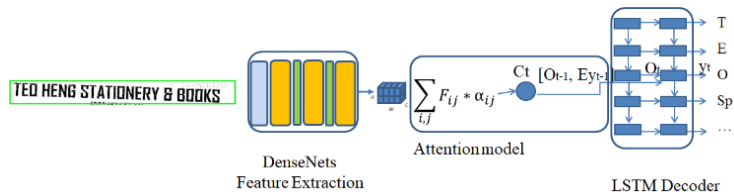

Fig. 3. Structure of Attention based Encoder-Decoder for text recognition.

(4) Finally, using the graph-based text line construction algorithm, the obtained one text segments are merged into a text line.

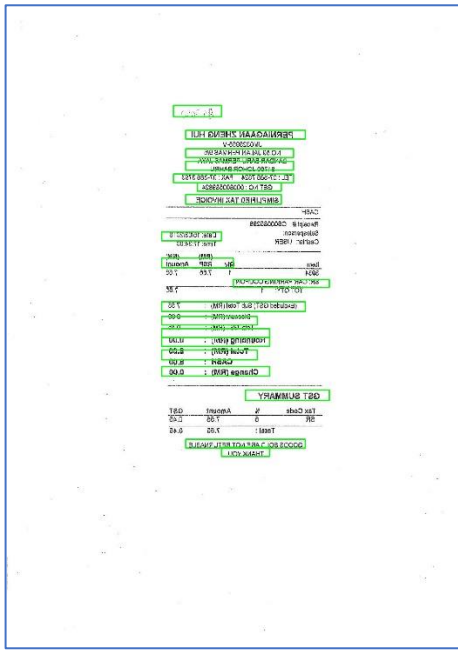

Fig. 4. A bad result of CTPN on a small receipt.

CTPN achieved a good result for horizontal text. However, CTPN cannot detect small text line, since VGG16 reduces 16 times of input images. Figure 4 shown a result of CTPN on a small receipt. Several text lines were not detected. To overcome this problem, we have to extract the receipt area from the input image. Therefore, the scale of the image is larger. We propose a simple method by the histograms of the input image on the x-axis and y-axis. The pre-process is as follows:

(1) We calculate the histogram of the input image on x-axis and y-axis.
(2) We determine the largest receipt area on each axis when the histogram is larger than a threshold. Then, we determine the receipt area.

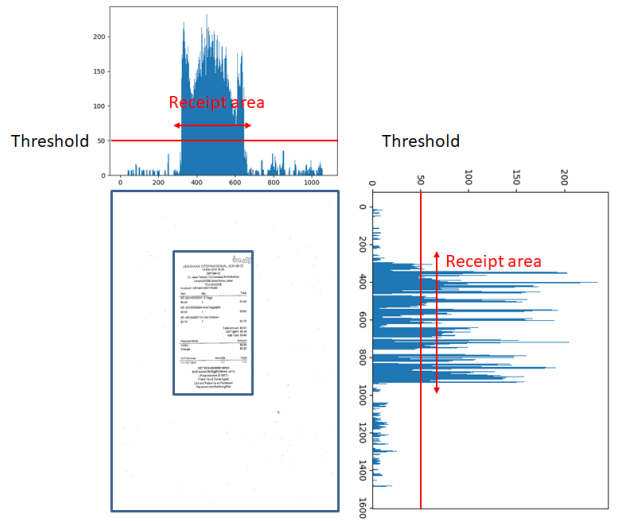

Fig. 5. Histograms on x-axis and y-axis of a receipt

## B. Text Recognition

The AED has been successfully applied for recognizing handwritten mathematical expression[4, 5], handwritten Vietnamese [6]. In this work, we employed the previous structure of ADE for recognizing text lines. The ADE has two main modules: DenseNet for extracting features from a text image and an LSTM combined with an attention model for predicting the output text. The process for recognition of a receipt is shown in Figure 4, and the detail of the AED is shown in Figure 3.

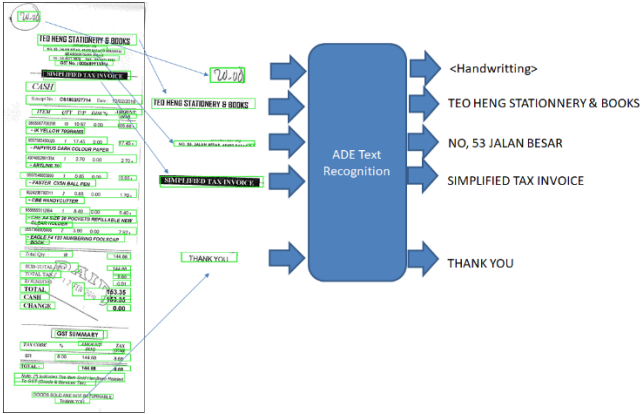

Fig. 6. Text recognition process.

**DenseNet Feature extraction:** Based on our previous researches on AED, we employed DenseNet as feature extraction. DenseNet has direct connections from any preceding layers to succeeding layers, so they help the network reuse and learn features cross layers. The detailed implementation was described in [4, 6]. We employed three dense blocks of growth rate (output feature map of each convolutional layer) $k = 24$ and the depth (number of convolutional layers in each dense block) $D = 16$ to extract features

**Attention-based LSTM decoder:** An LSTM decoder predicts one character at a time step. The decoder predicts the output symbol based on the embedding vector of the previous decoded symbol, the currently hidden state of the decoder, and the current context vector. The context vector is computed by the attention mechanism. The decoder is initialized by averaging the extracted features map.

**OCR verification:** We employ OCR to verify a text line as handwriting or printing. We added a category for handwriting on the decoder. For a handwriting text line, the ADE predicts as <handwriting> symbol. If a text line is recognized as <handwriting>, we will remove this text line from the result of text line detection.

## III. EVALUATION

### A. dataset

We employ the dataset from SROIE 2019 [7] for this research. Since the testing set of SROIE has not released, we divided the training data into training, validation, and testing sets. We randomly select 80% of receipts for training, 10% of receipts for validation, and the rest for testing. As a result, we have 500 receipts for training, 63 receipts for validation, and 63 for testing. For training text detection, we fine-tune a pre-trained network [8] for 50k iterations. For training text recognition, we extract text lines from receipts. The number of categories is 71, which contains many character categories such as Latin characters and numbers. To do OCR verification, we added a category for handwriting. We manually extracted 158 text lines for training and 23 text lines for validation from results of CTPN. We used mini-batch stochastic gradient descent to learn the parameters and early stopping for selecting the best model on the validation set.

### B. Evaluation metric

For text detection, we employed Tightness-aware Intersection-over-Union (TIoU) Metric for evaluation, which has some improvements: completeness of ground truth, compactness of detection, and tightness of matching degree. The harmonic mean of recall and precision is calculated as follows:

$$\text{Hmean} = 2 \frac{Recall * Precision}{Recall + Precision}$$
$$Recall = \frac{\sum Match_{gt_i}}{Num_{gt}}$$
$$Precision = \frac{\sum Match_{dt_i}}{Num_{dt}}$$

where $Match_{gt_i}$ and $Match_{dt_i}$ indicate the result of the match between detection and ground truth rectangles under TIoU conditions.

For text recognition, we employed $F_1$, Precision, Recall as follows:

$$F1 = 2 \frac{Recall * Precision}{Recall + Precision}$$
$$Recall = \frac{Number\ of\ correct\ words}{Number\ of\ recognized\ words}$$
$$Precision = \frac{Number\ of\ correct\ words}{Number\ of\ ground\ truth\ words}$$

### C. Results

Table I shows the result of CTPN with three conditions: CTPN: employing CTPN on original images; pre-processing + CTPN: employing pre-processing to extract receipt area; pre-processing + CTPN + OCR verification: employing pre-processing and OCR verification. The CTPN with pre-processing achieved the best result on Recall and Hmean while CTPN with pre-processing and OCR verification achieved the best result on Precision. The OCR verification can remove handwriting, but in some case, it removes text. The reason is that we have a few of handwriting patterns to train the OCR.

Table II shows the result of AED on ground truth of bounding boxes. ADE achieved 86.1% of F1.

TABLE I. THE RESULTS OF TEX DETECTION.

| Method | Recall | Precision | Hmean |
|---|---|---|---|
| CTPN | 45.2 | 72.9 | 55.8 |
| pre-processing +CTPN | **55.9** | 75.1 | **64.1** |
| pre-processing +CTPN+OCR verification | 53.9 | **77.5** | 63.6 |

TABLE II. THE RESULTS OF THE AED TEXT RECOGNITION ON GROUND TRUTH BOUNDING BOXES.

| Method | Recall | Precision | F1 |
|---|---|---|---|
| Ground Truth | 87.6 | 84.7 | 86.1 |

Table III shows the result of ADE on the detection results of CTPN. The result on bounding boxes of CTPN with pre-processing and OCR verification achieved the best result on F1 and Precision while the result on bounding boxes of CTPN with pre-processing achieved the best result on Recall.

TABLE III. THE RESULTS OF TEX DETECTION.

| Method | Recall | Precision | F1 |
|---|---|---|---|
| CTPN | 60.8 | 67.4 | 63.9 |
| pre-processing + CTPN | **72.3** | 69.5 | 70.9 |
| pre-processing + CTPN + OCR verification | 71.3 | **72.5** | **71.9** |

Figure 7 shows an example of a good result from our system. We figured out that the system will work well if the text detection provides precise results.

In the future, we plan to do the following works to improve our system and compare with other systems that participated in SROIE 2019.

+ Analyze error of our system for further improvements of text detection and text recognition.

+ Improve OCR verification by using handwriting from IAM database.

+ Do experiments on the testing set of SROIE 2019 when the testing set released.

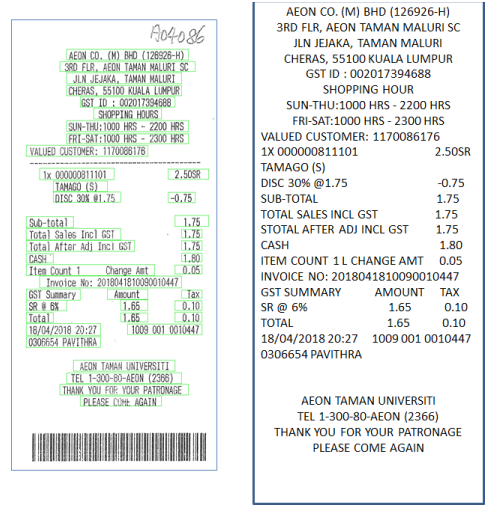

Fig. 7. An example of a good result by our recongition system.

## IV. CONCLUSION

In this paper, we have presented a deep learning based system for receipt recognition. We proposed receipt extraction and OCR verification to improve text detection. The system is able to recognize small receipt and ignore handwriting. We achieved 71.9% of F1 score for both detection and recognition task. This result is a baseline for the receipt recognition task.